\documentclass[letterpaper, 10 pt, conference]{ieeeconf}  % Comment this line out if you need a4paper
\usepackage{subfig}
\usepackage{hyperref}
\usepackage{todonotes}
\usepackage{graphicx}
\usepackage[nolist]{acronym}

\IEEEoverridecommandlockouts                              % This command is only needed if 
\title{\LARGE \bf
Scale-Robust Localization Using General Object Landmarks
}

\author{Andrew Holliday$^{1}$ and Gregory Dudek$^{2}$% <-this % stops a space
\thanks{$^{1}$Andrew Holliday and Gregory Dudek are with the Center for Intelligent Machines, McGill University, 845 Sherbrooke St, Montreal, Canada
        {\tt\small ahollid@cim.mcgill.ca, dudek@cim.mcgill.ca}}%
}

\begin{document}

\maketitle
\thispagestyle{empty}
\pagestyle{empty}

%%%%%%%%%%%%%%%%%%%%%%%%%%%%%%%%%%%%%%%%%%%%%%%%%%%%%%%%%%%%%%%%%%%%%%%%%%%%%%%%
\begin{abstract}

Visual localization under large changes in scale is an important capability in many robotic mapping applications, such as localizing at low altitudes in maps built at high altitudes, or performing loop closure over long distances.  Existing approaches, however, are robust only up to about a 3x difference in scale between map and query images.  

We propose a novel combination of deep-learning-based object features and state-of-the-art SIFT point-features that yields improved robustness to scale change.  This technique is training-free and class-agnostic, and in principle can be deployed in any environment out-of-the-box.  We evaluate the proposed technique on the KITTI Odometry benchmark and on a novel dataset of outdoor images exhibiting changes in visual scale of $7\times$ and greater, which we have released to the public.  Our technique consistently outperforms localization using either SIFT features or the proposed object features alone, achieving both greater accuracy and much lower failure rates under large changes in scale.

% We propose a training-free system for metric localization across large discrepancies in scale in arbitrary environments, based on a class-agnostic, non-semantic object proposal scheme and the use of rich, hierarchical feature descriptors from a convolutional neural network to match object detections between images and restrict the search for point feature matches.  

\end{abstract}

%%%%%%%%%%%%%%%%%%%%%%%%%%%%%%%%%%%%%%%%%%%%%%%%%%%%%%%%%%%%%%%%%%%%%%%%%%%%%%%%
\section{INTRODUCTION}

In this work, we attempt to address the problem of performing metric localization in a known environment under extreme changes in visual scale.  Our localization approach is based on the identification of objects in the environment, and their use as landmarks.  By ``objects" we here mean physical entities which are distinct from their surroundings and have some consistent physical properties of structure and appearance.

Many robotic applications involve repeated traversals of a known environment over time.  In such applications, it is usually beneficial to first construct a map of the environment, which can then be used by a robot to navigate the environment in subsequent missions.  Surveying the environment from a very high altitude allows complete geographic coverage of the environment to be obtained by shorter, and thus more efficient, paths by the surveyor.  At the same time, a robot that makes use of this high-altitude map to localize may have mission parameters requiring it to operate at a much lower altitude.

\begin{figure}
    \centering{}%

    \begin{minipage}[t]{0.2\textwidth}%
        \subfloat{\centering{}\includegraphics[width=1\textwidth]{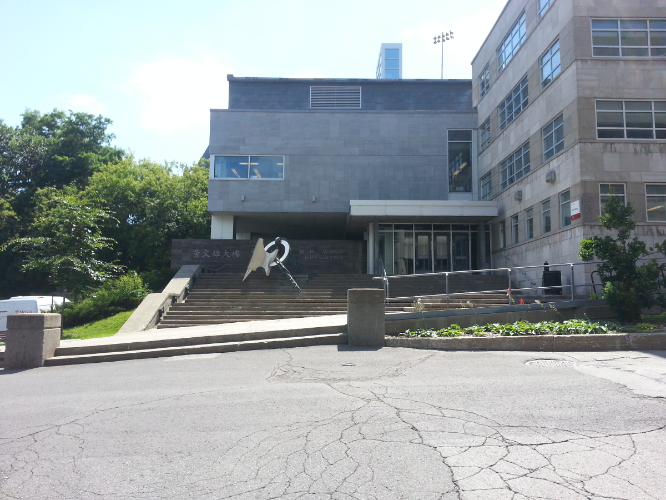}}%
    \end{minipage}\hspace{1px}
    \begin{minipage}[t]{0.2\textwidth}%
        \subfloat{\centering{}\includegraphics[width=1\textwidth]{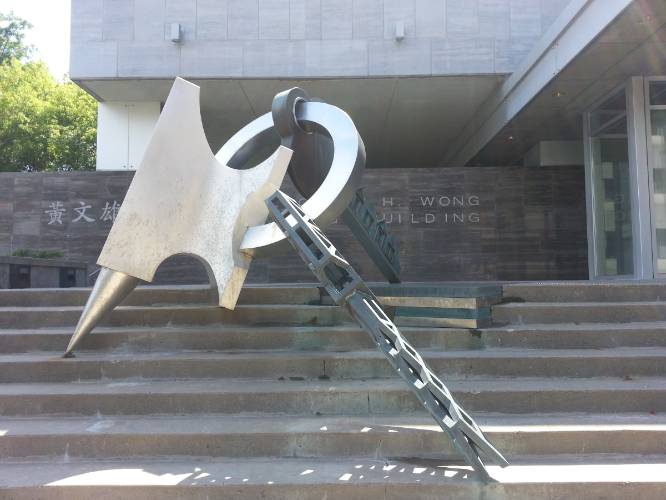}}%
    \end{minipage}\hspace{1px}\smallskip \\

    \begin{minipage}[t]{0.2\textwidth}%
        \begin{center}Far Image\par\end{center}%
    \end{minipage}\hspace{1px}
    \begin{minipage}[t]{0.2\textwidth}%
        \begin{center}Near Image\par\end{center}%
    \end{minipage}\smallskip \\

    \begin{minipage}[t]{0.2\textwidth}%
        \subfloat{\centering{}\includegraphics[width=1\textwidth]{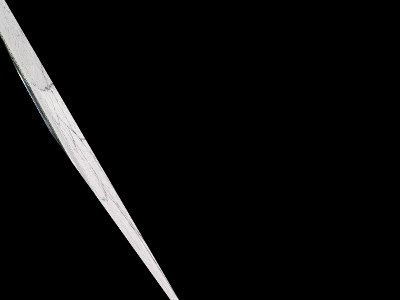}}
    \end{minipage}\hspace{1px}
    \begin{minipage}[t]{0.2\textwidth}%
        \subfloat{\centering{}\includegraphics[width=1\textwidth]{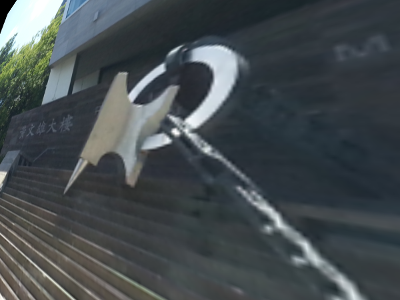}}
    \end{minipage}\smallskip \\
    
    \begin{minipage}[t]{0.2\textwidth}%
        \begin{center}SIFT\par\end{center}%
    \end{minipage}\hspace{1px}
    \begin{minipage}[t]{0.2\textwidth}%
        \begin{center}res3d, $64 \times 64$ input\par\end{center}%
    \end{minipage}\smallskip \\
    \caption{The results of SIFT features alone, and of the overall-best configuration of our system, on the largest-scale-change image pair in the dataset.  Despite the scale factor of 6, the highly-distinctive object in the foreground allows our system to determine an accurate homography between the images, while SIFT features fail to do so.}
    \label{fig:nearTrottier2}
\end{figure}

One such scenario is that of performing visual surveys of benthic environments, such as coral reefs, as in Johnson-Roberson et al.~\cite{underwaterMappingExample}.  A fast-moving surface vehicle may be used to rapidly map a large area of a reef.  This map may then be used by a slower-moving, but more maneuverable, autonomous underwater vehicle (AUV) such as the Aqua robot~\cite{aquaRobot}, to navigate the reef while capturing imagery very close to the sea floor.  Another relevant scenario is that of a robot performing loop closure over long distances as part of \ac{SLAM}.  Loop closure, the recognition of a previously-viewed location when viewing it a second time, is key to accurate \ac{SLAM}, and the overall accuracy of \ac{SLAM} techniques could be considerably improved if loop closure could be conducted across major changes in scale and perspective.

In scenarios such as the above, a robot must deal with the change in visual scale between two perspectives, which may be $5\times$ or even greater.  In some scenarios, such as in benthic environments, other factors may also intrude, such as colour-shifting due to the optical properties of water, and image noise due to particulate suspended in the water.  Identifying scenes across such large changes in scale is very challenging for modern visual localization techniques.  Even the most scale-robust techniques, such as \ac{SIFT}, can only localize reliably under scale factors less than about $3\times$.

We hypothesize that the hierarchical features computed by the intermediate layers of a \ac{CNN}~\cite{deepTextbook} may prove robust to changes in scale, due to their high degree of abstraction.  We propose a technique for performing metric localization across significant changes in scale by identifying and describing non-semantic objects in a way that allows them to be associated between scenes.  We show that these associations can be used to guide the matching of \ac{SIFT} features between images in a way that improves the robustness of matching to scale changes, allowing accurate localization under visual scale factors of 3 and greater.  The proposed system does not require any environment-specific training, and in principle can be deployed out-of-the-box in arbitrary environments.  The objects used by our system are defined functionally, in terms of their utility as scale-invariant landmarks, and are not limited to semantically-meaningful object categories.  

We specifically consider the problem of localizing between pairs of images known to contain the same scene at different visual scale.  A solution to this problem is an essential component of a system that can perform full global localization across large scale changes, and in certain cases - such as the low-vs-high-altitude case described above - could suffice on its own for global localization.  We demonstrate the approach both on standard localization benchmarks and on a novel dataset of image pairs from urban scenes exhibiting major scale changes.

\section{RELATED WORK}

Visual localization refers to the problem of determining a robot's pose using images from one or more cameras, with reference to a map or set of previously-seen images.  This may be done with some prior on the robot's position, or with no such prior, called global localization~\cite{dudek2010computational}.  Visual odometry is a form of non-global localization, while global localization is closely related to loop closure; both of these are important components of \ac{SLAM}, and there is a large body of literature exploring both problems.  Prominent early work includes Leonard et al.~\cite{leonard1991mobile} and Mackenzie et al.~\cite{mackenzie1994precise}, and Fox et al.~\cite{fox1999markov}.

Many traditional visual approaches to these problems, and particularly global localization, have been based on the recognition of whole-image descriptors of particular scenes, such as GIST features~\cite{gist}.  Successful instances include SeqSLAM~\cite{seqslam}, which uses a heavily downsampled version of the input image as a descriptor, and LSD-SLAM~\cite{lsdslam}, which performs direct image alignment for loop closure, as well as Hansen et al.~\cite{seqslamvar1}, Cadena et al.,~\cite{robustPlaceRecogWithSequences}, Liu et al.~\cite{zhangWholeImageComparison} and Naseer et al.~\cite{seqslamvar2}.  Because whole-image descriptors encode the geometric relationships of features in the 2D image plane, images of the same scene from different perspectives can have very different descriptors, making such methods very sensitive to changes in perspective and scale.

Another common approach is to discretize point-feature descriptors and build bag-of-words histograms of the input images.  FAB-MAP~\cite{fabmap}, ORB-SLAM~\cite{orbslam}, and the system of Ho et al.~\cite{seqslamvar3} perform variants of this for loop closure, starting from SURF~\cite{surf}, ORB~\cite{orb}, and \ac{SIFT}~\cite{sift} features, respectively.  While suitable for place-recognition tasks, such approaches alone are not appropriate for global localization, because spatial information about the visual words is not contained in the histogram.  
Hence, state-of-the-art \ac{SLAM} systems such as ORB-SLAM and LSD-SLAM rely on visual odometry for pose estimation.  Their visual odometry techniques are limited in robustness to changes in scale, perspective, and appearance, and so rely on successive estimations from closely-spaced frames.

Other global localization approaches attempt to recognize particular landmarks in an image, and use those to produce a metric estimate of the robot's pose.  SLAM++ of Salas-Moreno et al.~\cite{salas2013slam++} performs \ac{SLAM} by recognizing landmarks from a database of 3D object models.  Linegar et al.~\cite{landmarks24Hour} and Li et al.~\cite{highLevelFeaturesUnderwater} both train a bank of support vector machines (SVMs) to detect specific landmarks in a known environment, one SVM per landmark.  More recently,~\cite{semanticSlam} made use of a Deformable Parts Model (DPM)~\cite{deformablePartsModel} to detect objects for use as loop-closure landmarks in their \ac{SLAM} system.  All of these approaches require a pre-existing database of either object types or specific objects to operate.  These databases can be costly to construct, and these systems will fail in environments in which not enough landmarks belonging to the database are present.

Some work has explored the use of \acp{CNN} for localization.  PoseNet~\cite{posenet} is a \ac{CNN} that learns a mapping from images in an environment to metric camera poses, but it can only operate on the environment on which it was trained.  In S{\"{u}}nderhauf et al.~\cite{convWholeImagePlaceRecog}, the intermediate activations of a \ac{CNN} trained for image classification were used as whole-image descriptors for place recognition, a non-metric form of global localization.  In a similar fashion, Vysotska et al.~\cite{vysotska2016lazy} use whole-image descriptors from a \ac{CNN} in a SeqSLAM-like framework.  Subsequent work of S{\"{u}}nderhauf et al.~\cite{convNetLandmarks} refined this approach by using the same descriptor for object proposals within an image instead of the whole image.  Cascianelli et al.~\cite{cascianelli2017robust} and Panphattarasap et al.~\cite{panphattarasap2016visual} both expand on this technique.  These works consider only place recognition, however, and do not attempt to deal with the more challenging problem of full global localization (which necessitates returning a pose estimate).  Schmidt et al.~\cite{denseDeepPointDescriptors} and Simo-Serra et al.~\cite{sparseDeepPointDescriptors} have both explored the idea of learning point-feature descriptors with a \ac{CNN}, which could replace classical point features in a bag-of-words model.

When exploring robustness to perspective change, all of these works only consider positional variations of at most a few meters, when the scenes exhibit within-image scale variations of tens or hundreds of meters, and when the reference or training datasets consisted of images taken over traversals of environments ranging from hundreds to thousands of meters.  As a result, little significant change in scale exists between map images and query images in these experiments.  To the best of our knowledge, ours is the first to attempt to combine deep object-like features and point features into a single, unified representation of landmarks.  This synthesis provides superior metric localization to either technique in isolation, particularly under significant ($3\times$ and greater) changes in scale.

\section{PROPOSED SYSTEM}\label{sec:system}

The first stage of our metric localization pipeline consists in detecting objects in a pair of images, computing convolutional descriptors for them, and matching these descriptors between images. Our approach here closely follows that used for image-retrieval by  S{\"{u}}nderhauf et al.~\cite{convWholeImagePlaceRecog}; we differ in using Selective Search (SS), as proposed by Uijlings et al.~\cite{selectiveSearch}, to propose object regions, and in our use of a more recent \ac{CNN} architecture.

To extract objects from an image, Selective Search object proposals are first extracted from the image, and filtered to remove objects with bounding boxes less than 200 pixels in size and with aspect ratio greater than 3 or less than 1/3.  The image regions defined by each surviving SS bounding box are then extracted from the image, rescaled to a fixed size via bilinear interpolation, and run through a \ac{CNN}.  We use a ResNet-50 architecture trained on the ImageNet image-classification dataset, as described in He et al.~\cite{resnet}.  Experiments were run using six different layers of the network as feature descriptors, and with inputs to the network of four different resolutions.  The network layers and resolutions are listed in Table \ref{table:resNetSizes}.

Having extracted objects and their descriptors from a pair of images, we perform brute-force matching of the objects between the images.   Following~\cite{convNetLandmarks}, we take the match of each object descriptor $\mathbf{u}$ in image $i$ to be the descriptor $\mathbf{v}$ in image $j$ that has the smallest cosine distance from $\mathbf{u}$, defined as $t_{cos. err.}(\mathbf{u}, \mathbf{v}) = 1 - \frac{\mathbf{u} \cdot \mathbf{v}}{||\mathbf{u}||_2 \cdot ||\mathbf{v}||_2}$.  Matches are validated by cross-checking; a match $(\mathbf{u}, \mathbf{v})$ is only considered valid if $\mathbf{u}$ is the most similar object to $\mathbf{v}$ in image $i$ and $\mathbf{v}$ is the most similar object to $\mathbf{u}$ in image $j$.

\begin{table}[ht]
\caption{The sizes, as a number of floating-point values, of the output layers of ResNet-50 at different input resolutions.  Values in bold indicate layer-resolution pairs which provided the best results in any of our experiments.}
\begin{center}
\begin{tabular}{c|c c c c c c}
    Input res. & pool1 & res2c & res3d & res4f & res5c & pool5 \\
    \hline
    $224 \times 224$ & 201k & 803k & 401k & 200k & \textbf{100k} & 2k \\
    $128 \times 128$ & 66k & 262k & 131k & 65k & 131k & 8k \\
    $64 \times 64$ & 16k & 66k & \textbf{32k} & 66k & \textbf{131k} & 8k \\
    $32 \times 32$ & 4k & 16k & 32k & 66k & 131k & 8k \\
\end{tabular}
\end{center}
\label{table:resNetSizes}
\end{table}

% \begin{equation}\label{eqn:cosineDist}
% t_{cos\_err}(\mathbf{u}, \mathbf{v}) = 1 - \frac{\mathbf{u} \cdot \mathbf{v}}{||\mathbf{u}||_2 \cdot ||\mathbf{v}||_2}
% \end{equation}

Once object matches are found, we extract \ac{SIFT} features from both images, using 3 octave layers, an initial Gaussian with $\sigma=1.6$, an edge threshold of 10, and a contrast threshold of 0.04.  For each pair of matched objects, we match \ac{SIFT} features that lie inside the corresponding bounding boxes to one another.  \ac{SIFT} features are matched via their Euclidean distance, and cross-checking is again used to filter out bad matches.  By limiting the space over which we search for \ac{SIFT} matches to matched object regions, we hypothesize that the scope for error in \ac{SIFT} matching will be significantly reduced, and thus the accuracy of the resulting metric pose estimates will be increased.  As a baseline against which to compare our results, experiments were also run using \ac{SIFT} alone, with no objects, and objects alone, without \ac{SIFT} features - this last is essentially a na\"ive application of the place-recognition system of S{\"{u}}nderhauf~\cite{convNetLandmarks} to metric localization.  In these baseline experiments, \ac{SIFT} matching was performed in the same way, but the search for matches was conducted over all \ac{SIFT} features in both images.  When object proposals alone were used, they were matched in the same manner described above, and their bounding box centers were used as match points.

The resulting set of match points are used to produce a metric pose estimate.  Depending on the experiment, we compute either a homography $H$ or an essential matrix $E$~\cite{Hartley2004}.  In either case, the calculation of $H$ or $E$ from point correspondences is done via a RANSAC algorithm with an inlier threshold of 6, measured in pixel units.

\begin{figure}
    \centering
    \includegraphics[width=8.5cm,keepaspectratio]{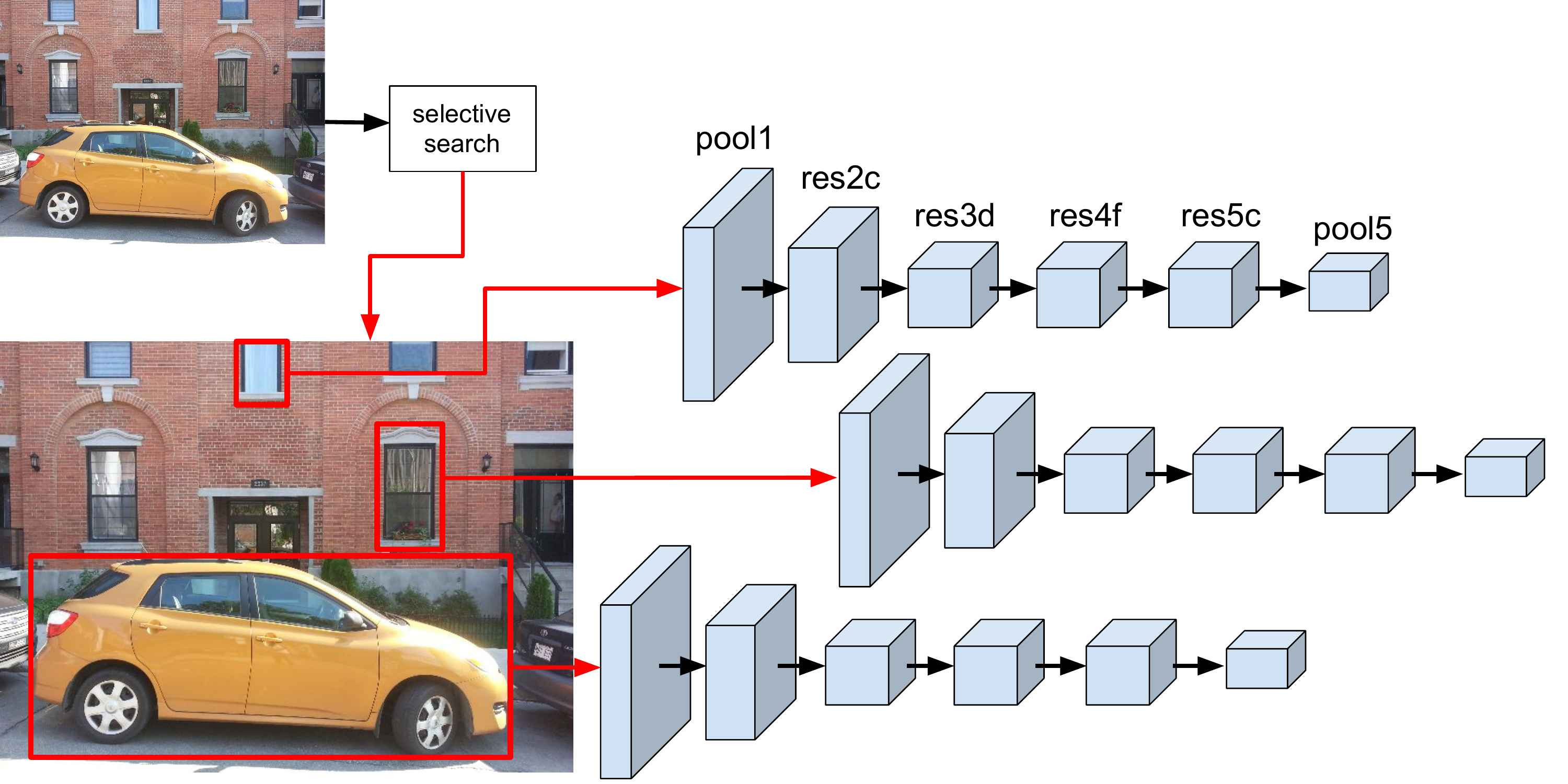}
    \caption{A simplified illustration of our object-detection architecture.}
    \label{fig:ssResNetArch}
\end{figure}

\section{KITTI EXPERIMENTS}\label{sec:kitti}

\subsection{Experimental Setup}

To evaluate the robustness of our proposed method to changes in scale, we conducted experiments on the KITTI Odometry benchmark dataset~\cite{kitti}.  This dataset consists of data sequences from a variety of sensors, including colour stereo imagery captured at a 15Hz frame rate, taken from a sensor rig mounted on a car as it drives along twenty-two distinct routes in the daytime.  Eleven of these sequences contain precise ground truth poses for each camera frame taken on each trajectory.  These trajectories were used to evaluate the proposed method.

\begin{figure}
    \centering
    \includegraphics[width=6cm]{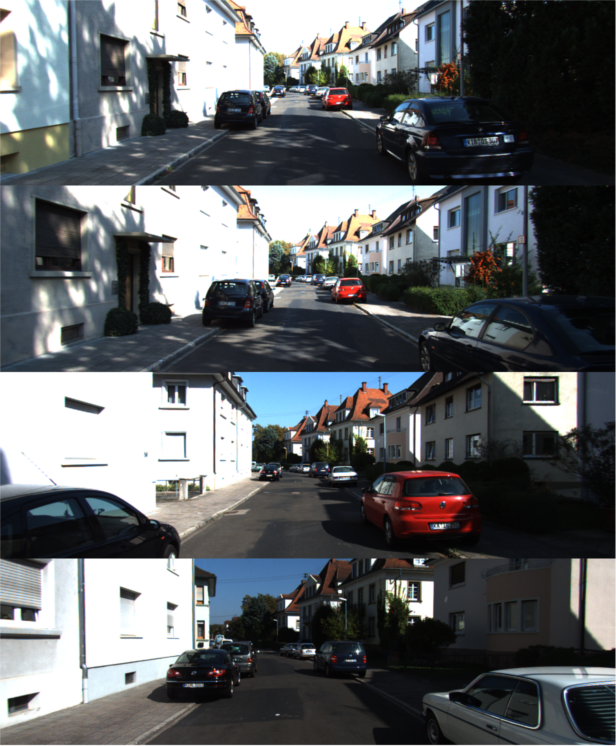}
    \caption{Sample frame from sequence 00.  The frames below it are separated by $j = 1, 5, 10$ in the subsampled sequence, respectively.  This gives an indication of the range of visual scale changes observed over all the pairs of the dataset.}
    \label{fig:kittiScalePairSamples}
\end{figure}

Our evaluation consisted of first subsampling each sequence by taking every fifth frame, to make the size of the overall dataset more manageable and increase the level of scale change present between adjacent frames in the sequence.  A set of image pairs was generated for each subsampled sequence by taking each frame $i$ in the sequence and pairing $i$ with the 10 subsequent frames, $i + j\ \forall j \in [1, 10]$.  Each successive value of $j$ gave an image pair $(i, i+j)$ with a greater degree of visual scale change, as shown in Fig.~\ref{fig:kittiScalePairSamples}.  

We finally filtered out any frame pairs whose gaze directions differed by more than $45^\circ$ in any axis, in order to consider only pairs that actually look at the same scene (in practice, only the yaw differs significantly in KITTI).  In total, 40,748 image pairs were used in our evaluation.  For each image pair, the images from the left colour camera (designated camera 2 in KITTI) were used for localization.  An example set of images is shown in Fig.~\ref{fig:kittiScalePairSamples}.

To estimate a transform between an image pair, a set of point matches was produced between the two images according to each of the three methods we compare, as described in section~\ref{sec:system}.  In each case, these point matches were used to estimate an essential matrix $E$, from which was derived a pose estimate $(\mathbf{q}_e, \mathbf{t}_e)$ via a standard method of applying SVD and cheirality checking~\cite{Hartley2004}.  $(\mathbf{q}_e, \mathbf{t}_e)$ describes the transform between the two frames.  To assess the quality of the estimate, we used two error metrics.  The first was the relative positional error, as defined in Eq.~\ref{eqn:relDist}:

\begin{equation}\label{eqn:relDist}
t_{err} = \frac{||\mathbf{t}_g - \mathbf{t}_e||_2}{||\mathbf{t}_g||_2 + ||\mathbf{t}_e||_2}
\end{equation}

where $t_g$ is the ground-truth translation between the two frames and $t_e$ is the estimated translation.  We normalize the vector from the estimated pose to the true pose to remove any correlation of that vector's length with the magnitude of the true translation.  Values of $t_{err}$ range from 0 to 1.

The second error metric was the rotational error, which following~\cite{rotationMetrics} is defined in Eq.~\ref{eqn:rotDist}:

\begin{equation}\label{eqn:rotDist}
r_{err} = 1 - |\mathbf{q}_g \cdot \mathbf{q}_e|
\end{equation}

Where $\mathbf{q}_g$ and $\mathbf{q}_e$ are quaternions representing the ground-truth and estimated gaze directions, respectively.  For some image pairs, no pose could be estimated, due to insufficient or inconsistent point matches.  We refer to this as localization failure, and for both $t_{err}$ and $r_{err}$ we substitute a value of 1, the maximum possible error under each metric, in these failure cases.

A preliminary evaluation was carried out over the space of CNN input resolutions and output layers by running them on the first 1000 image pairs from the first subsampled sequence (sequence 00).  We found that using an input resolution of $224 \times 224$ and the res5c feature layer as output gave both the highest accuracy and lowest localization failure rate.  This configuration was used for all object-landmark experiments on KITTI that we describe below.

\subsection{Results}\label{subsec:kittiResults}

\begin{figure}
    \centering
    \includegraphics[width=7.5cm,keepaspectratio]{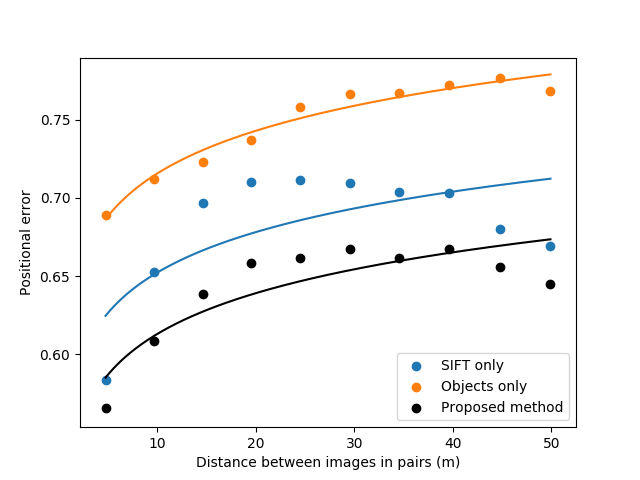}
    \caption{This plot shows the mean normalized positional error ($t_{err}$) versus the mean inter-camera distance of each group of image pairs.  Image pairs are grouped by the number of frames separating them in the sequence, from 1 to 10.  $t_{err}$ is a unitless error metric that ranges from 0 (best) to 1 (worst).  The improvement of the proposed method over SIFT is small but consistent.}
    \label{fig:kittiPosErr}
\end{figure}

\begin{figure}
    \centering
    \includegraphics[width=7.5cm,keepaspectratio]{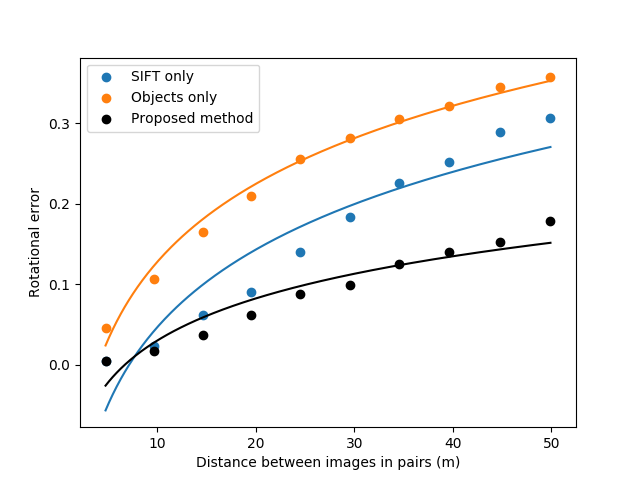}
    \caption{The mean rotational error ($r_{err}$) versus the mean inter-camera distance of each group of image pairs.  ($r_{err}$), like ($t_{err}$), is a unitless error metric that ranges from 0 (best) to 1 (worst).  This metric shows a more significant improvement over SIFT in the proposed method.}
    \label{fig:kittiRotErr}
\end{figure}

\begin{figure}
    \centering
    \includegraphics[width=7.5cm,keepaspectratio]{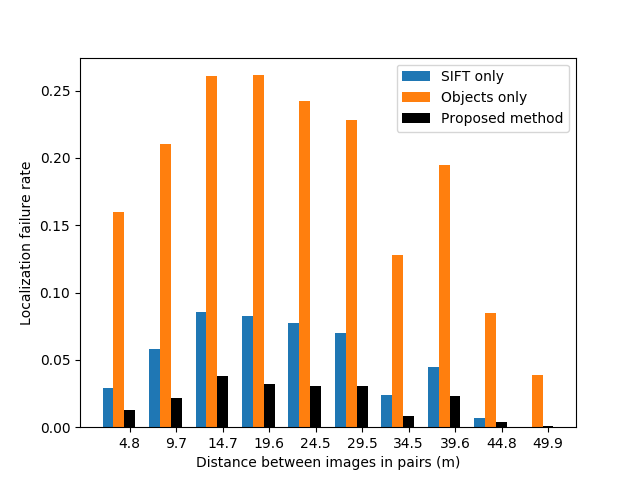}
    \caption{The fraction of the pairs in each image-pair group for which localization failure occurred, versus the mean inter-camera distance of each group.  In all groups, our proposed method has far fewer failures than either SIFT or object features alone.}
    \label{fig:kittiFailureRates}
\end{figure}

All metrics were plotted against the ground-truth translational distance, $||\mathbf{t}_g||_2$, between the frames in the image pairs.  To make these plots readable, we grouped image pairs by their frame-separation $j$, and plotted the mean error of each group against its mean ground-truth distance, in Fig.~\ref{fig:kittiPosErr} (for $t_{err}$) and Fig.~\ref{fig:kittiRotErr} (for $r_{err}$).  A logarithmic curve was fitted against each, as we expected that performance would initially worsen rapidly with distance, then level off.  We also display the failure rate of each group versus the group's mean distance in Fig.~\ref{fig:kittiFailureRates}.

\begin{table}[ht]
\caption{}
\label{table:kittiPerformance}
\begin{center}
\begin{tabular}{c|c c c }
Method & $t_{err}$ & $r_{err}$ & failure count \\
\hline
SIFT only & 0.680 & 0.149 & 1854 \\
Objects only & 0.744 & 0.232 & 7146 \\
Proposed method & \bf{0.641} & \bf{0.086} & \bf{785} \\
\end{tabular}
\end{center}
\end{table}

The overall performance of each method across all pairs is provided in Table~\ref{table:kittiPerformance}.  This table shows that our proposed method improves on SIFT under each metric: a small improvement of 6\% in $t_{err}$, and more significant improvements of 43\% in $r_{err}$, and 58\% in failure rate, overall.   Meanwhile, the objects-only method performs significantly worse than both our method and SIFT on all metrics and at all pair distances.  

Fig.~\ref{fig:kittiRotErr} shows that on $r_{err}$ the improvement of our method over SIFT is negligible at $j=1$, but grows significantly and consistently with the distance between frames.  In Fig.~\ref{fig:kittiPosErr} meanwhile, we see that on $t_{err}$ the improvement grows at first, and is greater than 0.05 for most of the intermediate gaps, but shrinks again at the largest gaps.  Fig.~\ref{fig:kittiFailureRates} shows similar behaviour in the localization failure rate - it is lowest for all methods at the largest gaps.

From visual inspection of these extreme image pairs, this improvement at high $j$ appears to be caused by sections where the vehicle drives down a long, straight road for some distance.  In these cases, the visual scale of objects visible near the end of the road will show little change over even a gap of $j = 10$, making localization relatively easy.  Unlike more winding roads, such long, straight sections will not have any high-$j$ pairs removed due to the images being on either side of a bend in the road, meaning that the high-$j$ groups will contain disproportionately many pairs from these straight sections.

\section{MONTREAL IMAGE PAIR EXPERIMENTS}

\subsection{Experimental Setup}

To test the effectiveness of the proposed system in a real-world scenario, a set of 31 image pairs were taken across eleven scenes surrounding the Montreal campus.  Scenes were chosen to contain a roughly-centred object of approximately uniform depth in the scene, so that a uniform change in image scale could be achieved by taking images at various distances from the object.  This ensures that successful matches must be made under one change in scale, and makes the relationship between the images amenable to description by a homography.  The image pairs exhibit changes in scale ranging from factors of about 1.5 to about 7, with the exception of one image pair showing scale change of about 15 in a prominent foreground object.  All images were taken using the rear-facing camera of a Samsung Galaxy S3 phone, and were downsampled to $1200\times900$ pixels via bilinear interpolation for all experiments.  Each image pair was hand-annotated with a set of 10 point correspondences, distributed approximately evenly over the nearer image in each pair.  We have made this dataset publicly available \footnote{\url{http://www.cim.mcgill.ca/~mrl/montreal\_scale\_pairs/}}.

The proposed system was used to compute point matches between each image pair, and from these point matches, a homography $H$ was computed as described in section~\ref{sec:system}.  $H$ was used to calculate the total symmetric transfer error (STE) for the image pair $e$ over the ground truth points:

\begin{equation}
    \textup{STE} = \sum^N_i||p^i_{far} - H p^i_{near}||_2 + ||p^i_{near} - H^{-1}p^i_{far}||_2
\end{equation}

Whenever no $H$ could be found for an image pair by some method, its error on that image pair was set to the maximum STE we observed for any attempted method, $\textup{STE}_\textup{max} = 15,833,861,380.8$. The plain STE ranges over many orders of magnitude on this dataset, so we present the results using the logarithmic STE, making the results easier to interpret.  

The same set of parameters were run over this dataset as in the KITTI experiments - our system at six network layers and four input resolutions, plus \ac{SIFT} alone and objects alone, for comparison.  However, the results from objects alone were substantially worse at all configurations than those of either SIFT or the proposed method, similar to what we observed in section~\ref{sec:kitti}. For the sake of brevity, we ignore the objects-only results in the discussion and figures below.

\subsection{Results}\label{subsec:realWorldResults}

Table~\ref{table:realErrorByLayer} shows the performance of each feature layer and each input resolution over the whole Montreal dataset, and shows the results from using \ac{SIFT} features alone as well.  As this table shows, the total error using just \ac{SIFT} features is significantly greater than that of the best-performing input resolution for each feature layer.  Also, the average error of the intermediate layers res2c, res3d, and res4f, are all very comparable.  It is interesting to note that in this experiment, more intermediate layers are favoured, while the KITTI experiments favoured the highest resolution and the second-deepest layer of the network.  This may arise from the difference in the native resolution of the images - KITTI's image resolutions vary from sequence to sequence, but are all close to $1230\times370$.

\begin{figure}
    \centering
    \includegraphics[width=9cm]{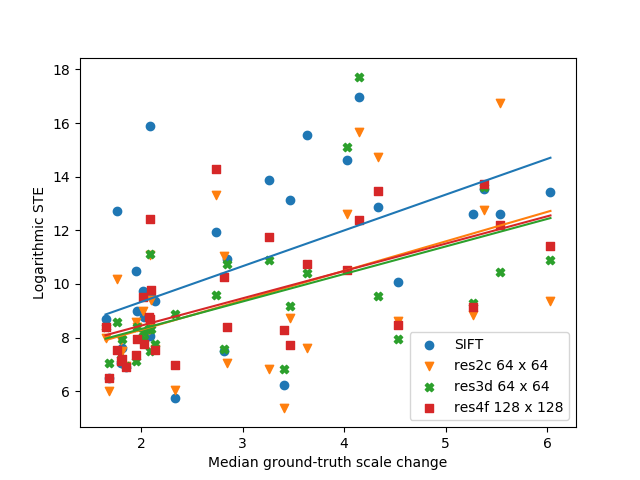}
    \caption{The error of \ac{SIFT} features alone, and the three best-performing configurations of our system, on each image pair in the dataset, plotted versus the median scale change exhibited in the image pair, along with a line of best fit for each method.}
    \label{fig:topThreeLayersPerPair}
\end{figure}

Fig.~\ref{fig:topThreeLayersPerPair} show the error of each of the three best-performing configurations, as well as the \ac{SIFT}-only approach, on each of the image pairs in the dataset, plotted versus the median scale change over all pairs of ground-truth matches $(p^i, p^j)$ in each image.  The scale change between matches $(p^i, p^j)$ is defined as: $\textup{scale change}_{i,j} = \frac{||p^i_{near} - p^j_{near}||_2}{||p^i_{far} - p^j_{far}||_2}$.  The lines of best fit for each method further emphasize the improvement of our system over \ac{SIFT} features at all scale factors up to 6.  The best-fit lines for all of the top-three configurations of our system overlap almost perfectly, although there is a fair degree of variance in their performances on individual examples.

% \begin{equation}\label{eqn:scaleChange}
%     \textup{scale change}_{i,j} = \frac{||p^i_{near} - p^j_{near}||_2}{||p^i_{far} - p^j_{far}||_2}
% \end{equation}

\begin{figure}
    \centering{}%

    \begin{minipage}[t]{0.12\textwidth}%
        \subfloat{\centering{}\includegraphics[width=1\textwidth]{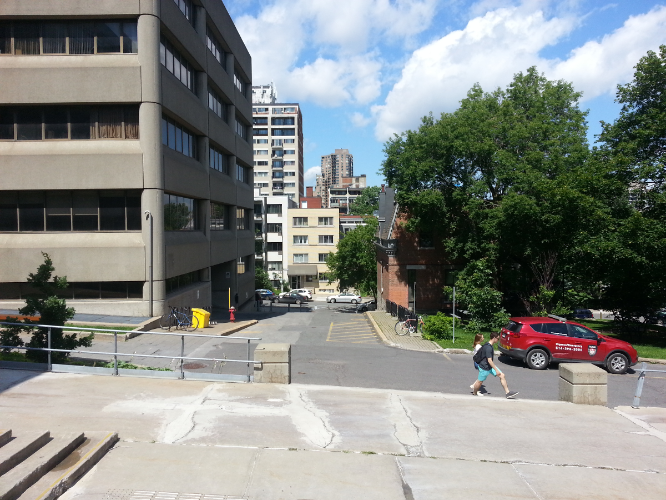}}%
    \end{minipage}\hspace{1px}
    \begin{minipage}[t]{0.12\textwidth}%
        \subfloat{\centering{}\includegraphics[width=1\textwidth]{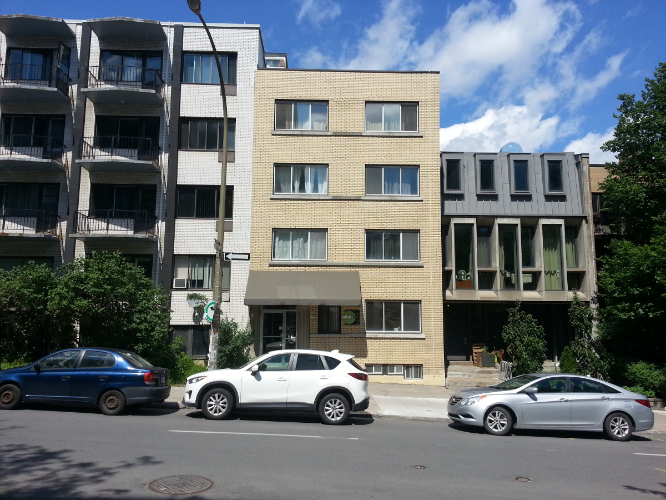}}%
    \end{minipage}\hspace{1px}
    \begin{minipage}[t]{0.12\textwidth}%
        \subfloat{\centering{}\includegraphics[width=1\textwidth]{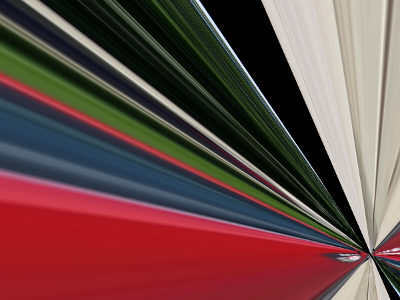}}
    \end{minipage}\smallskip \\

    \begin{minipage}[t]{0.12\textwidth}%
        \begin{center}Far Image\par\end{center}%
    \end{minipage}\hspace{1px}
    \begin{minipage}[t]{0.12\textwidth}%
        \begin{center}Near Image\par\end{center}%
    \end{minipage}\hspace{1px}
    \begin{minipage}[t]{0.12\textwidth}%
        \begin{center}SIFT\par\end{center}%
    \end{minipage}\smallskip \\
    
    \begin{minipage}[t]{0.12\textwidth}%
        \subfloat{\centering{}\includegraphics[width=1\textwidth]{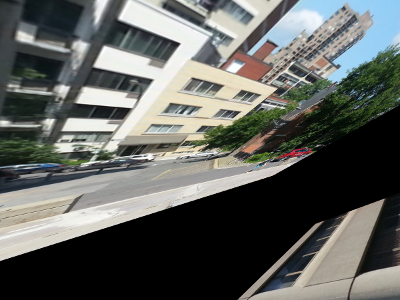}}%
    \end{minipage}\hspace{1px}
    \begin{minipage}[t]{0.12\textwidth}%
        \subfloat{\centering{}\includegraphics[width=1\textwidth]{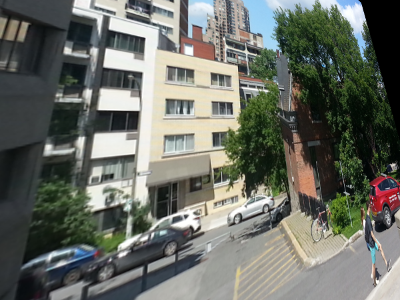}}%
    \end{minipage}\hspace{1px}
    \begin{minipage}[t]{0.12\textwidth}%
        \subfloat{\centering{}\includegraphics[width=1\textwidth]{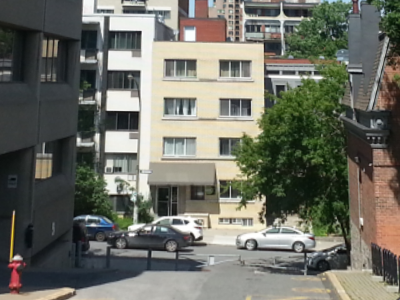}}
    \end{minipage}\smallskip \\
    
       \begin{minipage}[t]{0.12\textwidth}%
        \begin{center}res3d, $64 \times 64$ input\par\end{center}%
    \end{minipage}\hspace{1px}
    \begin{minipage}[t]{0.12\textwidth}%
        \begin{center}res4f, $128 \times 128$ input\par\end{center}%
    \end{minipage}\hspace{1px}
    \begin{minipage}[t]{0.12\textwidth}%
        \begin{center}res5c, $128 \times 128$ input\par\end{center}%
    \end{minipage}\smallskip \\
    \caption{The results of \ac{SIFT} features and three different configurations of our system on another pair.  The homography estimated by the best-overall configuration, res3d at $64 \times 64$ input size, is notably worse than those produced by two other intermediate feature layers.  No configuration of our system performs best on all image pairs.}
    \label{fig:nearTrotter1}
\end{figure}

The use of homographies to relate the image pairs allows us to visually inspect the quality of the estimated $H$, by using $H$ to map all pixels in the farther image to their estimated locations in the nearer image.  Visual inspection of these mappings for the 31 image pairs confirm that those configurations with lower logarithmic STEs tend to have more correct-looking mappings, although all configurations of our system with mean logarithmic STE $< 10$ produce comparable mappings for most pairs, and on some pairs, higher-error configurations such as res4f with $64 \times 64$-pixel inputs produce a subjectively better mapping than the lowest-error configuration.  Fig.~\ref{fig:nearTrottier2} and Fig.~\ref{fig:nearTrotter1} display some example homography mappings.

% \begin{figure}
%     \centering
%     \includegraphics[width=8cm,keepaspectratio]{Images/realWorld/realWorldBoxErrors.png}
%     \caption{A table showing the logarithmic STE of each configuration of the system, averaged over all image pairs.  The best-performing feature overall is res3d with a $64 \times 64$ input size, followed closely by res2c with $64 \times 64$ inputs and res4f with $128 \times 128$ inputs.}
%     \label{fig:realNetworkPerformanceTablePlot}
% \end{figure}

\begin{table}[ht]
\caption{A table showing the logarithmic STE of each configuration of the system, averaged over all image pairs.  The best-performing feature overall is res3d with a $64 \times 64$ input size, followed closely by res2c with $64 \times 64$ inputs and res4f with $128 \times 128$ inputs.  The mean log. STE of \ac{SIFT} features alone is presented as well, for comparison.}
\label{table:realErrorByLayer}
\begin{center}
\begin{tabular}{c|c c c c c c}
Input res. & pool1 & res2c & res3d & res4f & res5c & pool5 \\
\hline
$224 \times 224$ & 11.102 & 13.057 & 9.631 & 10.277 & 10.537 & 10.011 \\
$128 \times 128$ & 10.389 & 11.716 & 10.231 & 9.458 & 9.930 & 9.505 \\
$64 \times 64$ & 10.921 & 9.381 & \textbf{9.339} & 9.777 & 10.234 & 9.667 \\
$32 \times 32$ & 10.134 & 10.162 & 10.473 & 9.658 & 10.301 & 10.607 \\
\hline
SIFT & 10.654
\end{tabular}
\end{center}
\end{table}

\section{CONCLUSIONS}

\addtolength{\textheight}{-0.5cm}   % This command serves to balance the column lengths
                                  % on the last page of the document manually. It shortens
                                  % the textheight of the last page by a suitable amount.
                                  % This command does not take effect until the next page
                                  % so it should come on the page before the last. Make
                                  % sure that you do not shorten the textheight too much.

One strength of our proposed system is that it requires no domain-specific training, making use only of a pre-trained \ac{CNN}.  However, as future work we wish to explore the possibility of training a \ac{CNN} with the specific objective of producing a scale- and perspective-invariant object descriptor, as doing so may result in more accurate matching of objects.  We also wish to explore the possibility that including matches from multiple layers of the network in the localization process could improve the system's accuracy.

The most natural extension of this work, however, is to extend it to the full global-localization problem, where the system must localize within a large map or database of images with no prior on the position, and must moreover do so across major scale changes.  Depending on the scenario, this may require combining our localization method with a similarly scale-robust place-recognition system.

We have shown that by combining deep learning with classical methods, we can perform accurate localization across major changes in scale.  Our system uses a pre-trained deep network to describe arbitrary objects and correctly match them between images for use as navigation landmarks.  Restricting \ac{SIFT} feature matching to matched object regions substantially improves the robustness of \ac{SIFT} matching both to changes in image noise and to changes in scale.  Despite much prior work on place recognition and localization using both classical methods and deep learning, our result sets a new benchmark for metric localization performance across significant scale changes.

\begin{acronym}
\acro{CNN}{Convolutional Neural Network}
\acro{SLAM}{Simultaneous Localization And Mapping}
\acro{SIFT}{Scale-Invariant Feature Transforms}
\end{acronym}

\bibliographystyle{IEEEtran}
\bibliography{IEEEabrv,mybib}

\begin{thebibliography}{10}
\providecommand{\url}[1]{#1}
\csname url@rmstyle\endcsname
\providecommand{\newblock}{\relax}
\providecommand{\bibinfo}[2]{#2}
\providecommand\BIBentrySTDinterwordspacing{\spaceskip=0pt\relax}
\providecommand\BIBentryALTinterwordstretchfactor{4}
\providecommand\BIBentryALTinterwordspacing{\spaceskip=\fontdimen2\font plus
\BIBentryALTinterwordstretchfactor\fontdimen3\font minus
  \fontdimen4\font\relax}
\providecommand\BIBforeignlanguage[2]{{%
\expandafter\ifx\csname l@#1\endcsname\relax
\typeout{** WARNING: IEEEtran.bst: No hyphenation pattern has been}%
\typeout{** loaded for the language `#1'. Using the pattern for}%
\typeout{** the default language instead.}%
\else
\language=\csname l@#1\endcsname
\fi
#2}}

\bibitem{underwaterMappingExample}
M.~Johnson-Roberson, O.~Pizarro, S.~B. Williams, and I.~Mahon, ``Generation and
  visualization of large-scale three-dimensional reconstructions from
  underwater robotic surveys,'' \emph{Journal of Field Robotics}, vol.~27,
  no.~1, pp. 21--51, 2010.

\bibitem{aquaRobot}
J.~Sattar, G.~Dudek, O.~Chiu, I.~Rekleitis, P.~Gigu\`ere, A.~Mills,
  N.~Plamondon, C.~Prahacs, Y.~Girdhar, M.~Nahon, and J.-P. Lobos, ``Enabling
  autonomous capabilities in underwater robotics,'' in \emph{Proceedings of the
  IEEE/RSJ International Conference on Intelligent Robots and Systems, {IROS}},
  Nice, France, September 2008.

\bibitem{deepTextbook}
I.~Goodfellow, Y.~Bengio, and A.~Courville, ``Deep learning. book in
  preparation for mit press,'' \emph{URL< http://www. deeplearningbook. org},
  2016.

\bibitem{dudek2010computational}
G.~Dudek and M.~Jenkin, \emph{Computational principles of mobile
  robotics}.\hskip 1em plus 0.5em minus 0.4em\relax Cambridge university press,
  2010.

\bibitem{leonard1991mobile}
J.~J. Leonard and H.~F. Durrant-Whyte, ``Mobile robot localization by tracking
  geometric beacons,'' \emph{IEEE Transactions on robotics and Automation},
  vol.~7, no.~3, pp. 376--382, 1991.

\bibitem{mackenzie1994precise}
P.~MacKenzie and G.~Dudek, ``Precise positioning using model-based maps,'' in
  \emph{Robotics and Automation, 1994. Proceedings., 1994 IEEE International
  Conference on}.\hskip 1em plus 0.5em minus 0.4em\relax IEEE, 1994, pp.
  1615--1621.

\bibitem{fox1999markov}
D.~Fox, W.~Burgard, and S.~Thrun, ``Markov localization for mobile robots in
  dynamic environments,'' \emph{Journal of Artificial Intelligence Research},
  vol.~11, pp. 391--427, 1999.

\bibitem{gist}
\BIBentryALTinterwordspacing
A.~Oliva and A.~Torralba, ``Modeling the shape of the scene: A holistic
  representation of the spatial envelope,'' \emph{Int. J. Comput. Vision},
  vol.~42, no.~3, pp. 145--175, May 2001. [Online]. Available:
  \url{http://dx.doi.org/10.1023/A:1011139631724}
\BIBentrySTDinterwordspacing

\bibitem{seqslam}
\BIBentryALTinterwordspacing
M.~Milford and G.~Wyeth, ``Seqslam : visual route-based navigation for sunny
  summer days and stormy winter nights,'' in \emph{IEEE International Conferece
  on Robotics and Automation (ICRA 2012)}, N.~Papanikolopoulos, Ed.\hskip 1em
  plus 0.5em minus 0.4em\relax River Centre, Saint Paul, Minnesota: IEEE, 2012,
  pp. 1643--1649. [Online]. Available: \url{http://eprints.qut.edu.au/51538/}
\BIBentrySTDinterwordspacing

\bibitem{lsdslam}
J.~Engel, T.~Sch{\"o}ps, and D.~Cremers, ``Lsd-slam: Large-scale direct
  monocular slam,'' in \emph{European Conference on Computer Vision}.\hskip 1em
  plus 0.5em minus 0.4em\relax Springer, 2014, pp. 834--849.

\bibitem{seqslamvar1}
P.~Hansen and B.~Browning, ``Visual place recognition using hmm sequence
  matching,'' in \emph{2014 IEEE/RSJ International Conference on Intelligent
  Robots and Systems}, Sept 2014, pp. 4549--4555.

\bibitem{robustPlaceRecogWithSequences}
C.~Cadena, D.~Galvez-López, J.~D. Tardos, and J.~Neira, ``Robust place
  recognition with stereo sequences,'' \emph{IEEE Transactions on Robotics},
  vol.~28, no.~4, pp. 871--885, Aug 2012.

\bibitem{zhangWholeImageComparison}
Y.~Liu and H.~Zhang, ``Performance evaluation of whole-image descriptors in
  visual loop closure detection,'' in \emph{2013 IEEE International Conference
  on Information and Automation (ICIA)}, Aug 2013, pp. 716--722.

\bibitem{seqslamvar2}
\BIBentryALTinterwordspacing
T.~Naseer, L.~Spinello, W.~Burgard, and C.~Stachniss, ``Robust visual robot
  localization across seasons using network flows,'' in \emph{AAAI Conference
  on Artificial Intelligence}, 2014. [Online]. Available:
  \url{http://www.aaai.org/ocs/index.php/AAAI/AAAI14/paper/view/8483}
\BIBentrySTDinterwordspacing

\bibitem{fabmap}
M.~Cummins and P.~Newman, ``{Invited Applications Paper FAB-MAP:
  Appearance-Based Place Recognition and Mapping using a Learned Visual
  Vocabulary Model},'' in \emph{27th Intl Conf. on Machine Learning
  (ICML2010)}, 2010.

\bibitem{orbslam}
R.~Mur-Artal, J.~M.~M. Montiel, and J.~D. Tardós, ``Orb-slam: a versatile and
  accurate monocular slam system.'' \emph{CoRR}, vol. abs/1502.00956, 2015.

\bibitem{seqslamvar3}
\BIBentryALTinterwordspacing
K.~L. Ho and P.~Newman, ``Detecting loop closure with scene sequences,''
  \emph{Int. J. Comput. Vision}, vol.~74, no.~3, pp. 261--286, Sept. 2007.
  [Online]. Available: \url{http://dx.doi.org/10.1007/s11263-006-0020-1}
\BIBentrySTDinterwordspacing

\bibitem{surf}
\BIBentryALTinterwordspacing
H.~Bay, A.~Ess, T.~Tuytelaars, and L.~Van~Gool, ``Speeded-up robust features
  (surf),'' \emph{Comput. Vis. Image Underst.}, vol. 110, no.~3, pp. 346--359,
  June 2008. [Online]. Available:
  \url{http://dx.doi.org/10.1016/j.cviu.2007.09.014}
\BIBentrySTDinterwordspacing

\bibitem{orb}
\BIBentryALTinterwordspacing
E.~Rublee, V.~Rabaud, K.~Konolige, and G.~Bradski, ``Orb: An efficient
  alternative to sift or surf,'' in \emph{Proceedings of the 2011 International
  Conference on Computer Vision}, ser. ICCV '11.\hskip 1em plus 0.5em minus
  0.4em\relax Washington, DC, USA: IEEE Computer Society, 2011, pp. 2564--2571.
  [Online]. Available: \url{http://dx.doi.org/10.1109/ICCV.2011.6126544}
\BIBentrySTDinterwordspacing

\bibitem{sift}
\BIBentryALTinterwordspacing
D.~G. Lowe, ``Object recognition from local scale-invariant features,'' in
  \emph{Proceedings of the International Conference on Computer Vision-Volume 2
  - Volume 2}, ser. ICCV '99.\hskip 1em plus 0.5em minus 0.4em\relax
  Washington, DC, USA: IEEE Computer Society, 1999, pp. 1150--. [Online].
  Available: \url{http://dl.acm.org/citation.cfm?id=850924.851523}
\BIBentrySTDinterwordspacing

\bibitem{salas2013slam++}
R.~F. Salas-Moreno, R.~A. Newcombe, H.~Strasdat, P.~H. Kelly, and A.~J.
  Davison, ``Slam++: Simultaneous localisation and mapping at the level of
  objects,'' in \emph{Proceedings of the IEEE conference on computer vision and
  pattern recognition}, 2013, pp. 1352--1359.

\bibitem{landmarks24Hour}
C.~Linegar, W.~Churchill, and P.~Newman, ``Made to measure: Bespoke landmarks
  for 24-hour, all-weather localisation with a camera,'' in \emph{2016 IEEE
  International Conference on Robotics and Automation (ICRA)}, May 2016, pp.
  787--794.

\bibitem{highLevelFeaturesUnderwater}
R.~M.~E. Jie~Li and M.~Johnson-Roberson, ``High-level visual features for
  underwater place recognition.''

\bibitem{semanticSlam}
S.~L. Bowman, N.~Atanasov, K.~Daniilidis, and G.~J. Pappas, ``Probabilistic
  data association for semantic slam,'' in \emph{2017 IEEE International
  Conference on Robotics and Automation (ICRA)}, May 2017, pp. 1722--1729.

\bibitem{deformablePartsModel}
P.~Felzenszwalb, D.~McAllester, and D.~Ramanan, ``A discriminatively trained,
  multiscale, deformable part model,'' in \emph{Computer Vision and Pattern
  Recognition, 2008. CVPR 2008. IEEE Conference on}.\hskip 1em plus 0.5em minus
  0.4em\relax IEEE, 2008, pp. 1--8.

\bibitem{posenet}
A.~Kendall and R.~Cipolla, ``Modelling uncertainty in deep learning for camera
  relocalization,'' \emph{Proceedings of the International Conference on
  Robotics and Automation ({ICRA})}, 2016.

\bibitem{convWholeImagePlaceRecog}
\BIBentryALTinterwordspacing
N.~S{\"{u}}nderhauf, F.~Dayoub, S.~Shirazi, B.~Upcroft, and M.~Milford, ``On
  the performance of convnet features for place recognition,'' \emph{CoRR},
  vol. abs/1501.04158, 2015. [Online]. Available:
  \url{http://arxiv.org/abs/1501.04158}
\BIBentrySTDinterwordspacing

\bibitem{vysotska2016lazy}
O.~Vysotska and C.~Stachniss, ``Lazy data association for image sequences
  matching under substantial appearance changes,'' \emph{IEEE Robotics and
  Automation Letters}, vol.~1, no.~1, pp. 213--220, 2016.

\bibitem{convNetLandmarks}
A.~J. F. D. E. P. B.~U. Niko S~̈underhauf, Sareh~Shirazi and M.~Milford,
  ``Place recognition with convnet landmarks: Viewpoint-robust,
  condition-robust, training-free,'' in \emph{Proceedings of Robotics: Science
  and Systems (RSS)}, 2015.

\bibitem{cascianelli2017robust}
S.~Cascianelli, G.~Costante, E.~Bellocchio, P.~Valigi, M.~L. Fravolini, and
  T.~A. Ciarfuglia, ``Robust visual semi-semantic loop closure detection by a
  covisibility graph and cnn features,'' \emph{Robotics and Autonomous
  Systems}, vol.~92, pp. 53--65, 2017.

\bibitem{panphattarasap2016visual}
P.~Panphattarasap and A.~Calway, ``Visual place recognition using landmark
  distribution descriptors,'' in \emph{Asian Conference on Computer
  Vision}.\hskip 1em plus 0.5em minus 0.4em\relax Springer, 2016, pp. 487--502.

\bibitem{denseDeepPointDescriptors}
T.~Schmidt, R.~Newcombe, and D.~Fox, ``Self-supervised visual descriptor
  learning for dense correspondence,'' \emph{IEEE Robotics and Automation
  Letters}, vol.~2, no.~2, pp. 420--427, 2017.

\bibitem{sparseDeepPointDescriptors}
E.~Simo-Serra, E.~Trulls, L.~Ferraz, I.~Kokkinos, P.~Fua, and F.~Moreno-Noguer,
  ``Discriminative learning of deep convolutional feature point descriptors,''
  in \emph{Proceedings of the IEEE International Conference on Computer
  Vision}, 2015, pp. 118--126.

\bibitem{selectiveSearch}
\BIBentryALTinterwordspacing
J.~R.~R. Uijlings, K.~E.~A. van~de Sande, T.~Gevers, and A.~W.~M. Smeulders,
  ``Selective search for object recognition,'' \emph{International Journal of
  Computer Vision}, vol. 104, no.~2, pp. 154--171, 2013. [Online]. Available:
  \url{https://ivi.fnwi.uva.nl/isis/publications/2013/UijlingsIJCV2013}
\BIBentrySTDinterwordspacing

\bibitem{resnet}
K.~He, X.~Zhang, S.~Ren, and J.~Sun, ``Deep residual learning for image
  recognition,'' \emph{arXiv preprint arXiv:1512.03385}, 2015.

\bibitem{Hartley2004}
R.~I. Hartley and A.~Zisserman, \emph{Multiple View Geometry in Computer
  Vision}, 2nd~ed.\hskip 1em plus 0.5em minus 0.4em\relax Cambridge University
  Press, ISBN: 0521540518, 2004.

\bibitem{kitti}
A.~Geiger, P.~Lenz, and R.~Urtasun, ``Are we ready for autonomous driving? the
  kitti vision benchmark suite,'' in \emph{Conference on Computer Vision and
  Pattern Recognition (CVPR)}, 2012.

\bibitem{rotationMetrics}
D.~Q. Huynh, ``Metrics for 3d rotations: Comparison and analysis,''
  \emph{Journal of Mathematical Imaging and Vision}, vol.~35, no.~2, pp.
  155--164, 2009.

\end{thebibliography}

%%%%%%%%%%%%%%%%%%%%%%%%%%%%%%%%%%%%%%%%%%%%%%%%%%%%%%%%%%%%%%%%%%%%%%%%%%%%%%%%
% \section*{APPENDIX}

% Appendixes should appear before the acknowledgment.

% \section*{ACKNOWLEDGMENT}

% TODO What goes here???  Anything? Cut it?

\end{document}